\documentclass[journal]{IEEEtran}
\newcommand{\ignore}[1]{}
\pdfoutput=1

\usepackage[utf8]{inputenc}
\usepackage[T1]{fontenc}
\usepackage{amsmath,epsfig}
\usepackage{graphicx}
\usepackage{epstopdf}
\usepackage{bm}
\usepackage{amssymb}
\usepackage{cite}
\usepackage{array}
\usepackage{multirow}
\usepackage{color}
\newcommand{\R}[1]{\textcolor[rgb]{1,0,0}{#1}}
\usepackage{ragged2e}
\renewcommand{\raggedright}{\leftskip=0pt \rightskip=0pt plus 0cm}

\begin{document}
%
\title{End-to-End Image Super-Resolution via Deep and Shallow Convolutional Networks }
\author{Yifan Wang, Lijun Wang, Hongyu Wang, Peihua Li \\
Dalian University of Technology, China\\
\{wyfan523, wlj\}@mail.dlut.edu.cn, \{whyu, peihuali\}@dlut.edu.cn}

\maketitle


\begin{abstract}
One impressive advantage of convolutional neural networks (CNNs) is their ability to automatically learn feature representation from raw pixels, eliminating the need for hand-designed procedures. However, recent methods for single image super-resolution (SR) fail to maintain this advantage. They utilize CNNs in two decoupled steps, \emph{i.e.}, first upsampling the low resolution (LR) image to the high resolution (HR) size with hand-designed techniques (\emph{e.g.}, bicubic interpolation), and then applying CNNs on the upsampled LR image to reconstruct HR results. In this paper, we seek an alternative and propose a new image SR method, which jointly learns the feature extraction, upsampling and HR reconstruction modules, yielding a completely end-to-end trainable deep CNN.
As opposed to existing approaches, the proposed method conducts upsampling in the latent feature space with filters that are optimized for the task of image SR.
In addition, the HR reconstruction is performed in a multi-scale manner to simultaneously incorporate both short- and long-range contextual information, ensuring more accurate restoration of HR images. To facilitate network training, a new training approach is designed, which jointly trains the proposed deep network with a relatively shallow network, leading to faster convergence and more superior performance. The proposed method is extensively evaluated on widely adopted data sets and improves the performance of state-of-the-art methods with a considerable margin. Moreover, in-depth ablation studies are conducted to verify the contribution of different network designs to image SR, providing additional insights for future research.
\end{abstract}


\IEEEpeerreviewmaketitle

\section{Introduction}
\label{sec:introduction}
\begin{figure}[top]
\centering
\begin{tabular}{c}
\includegraphics[width=1.02 \linewidth]{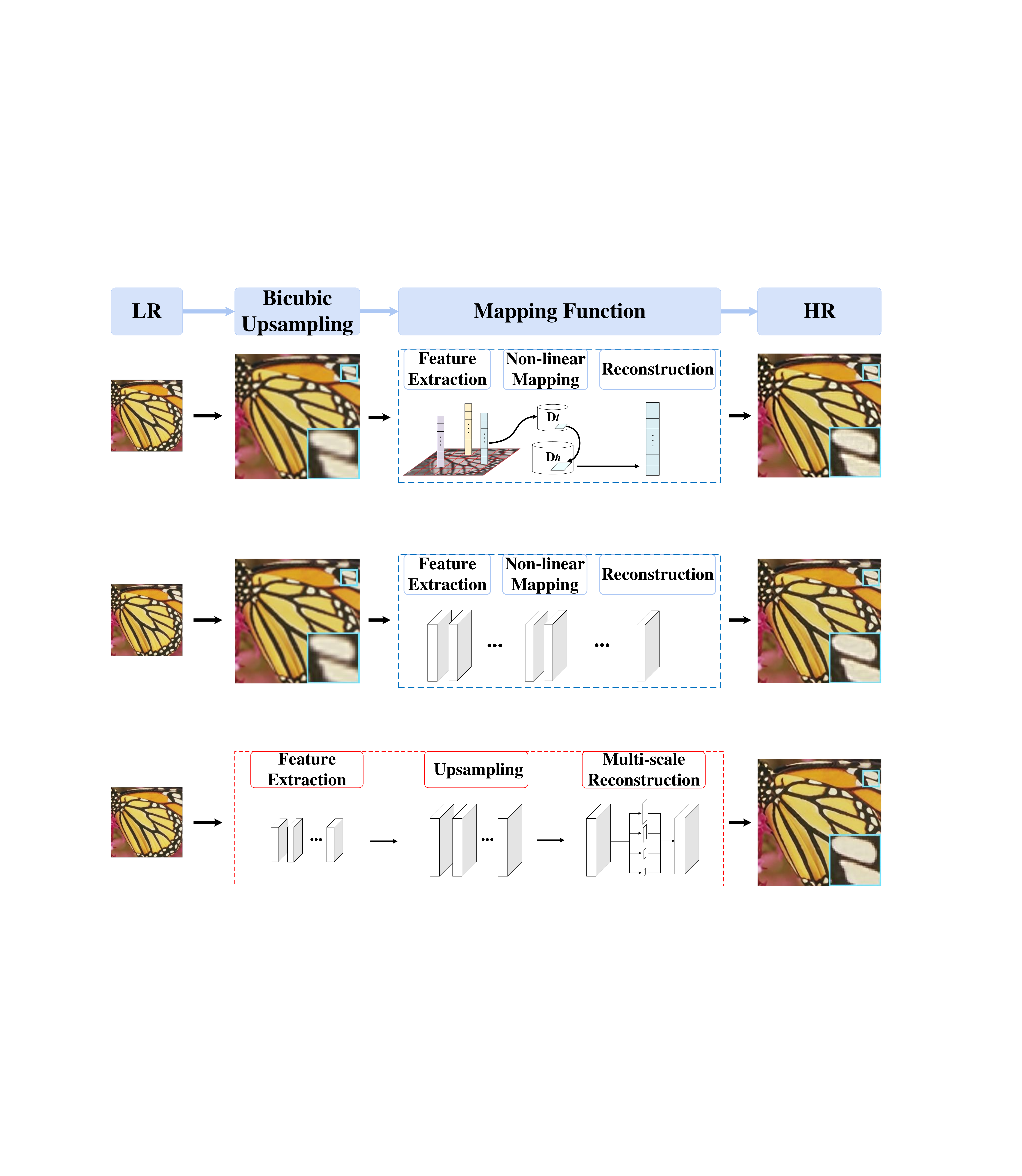}\\
\small{(a)}\\
\includegraphics[width=1.02 \linewidth]{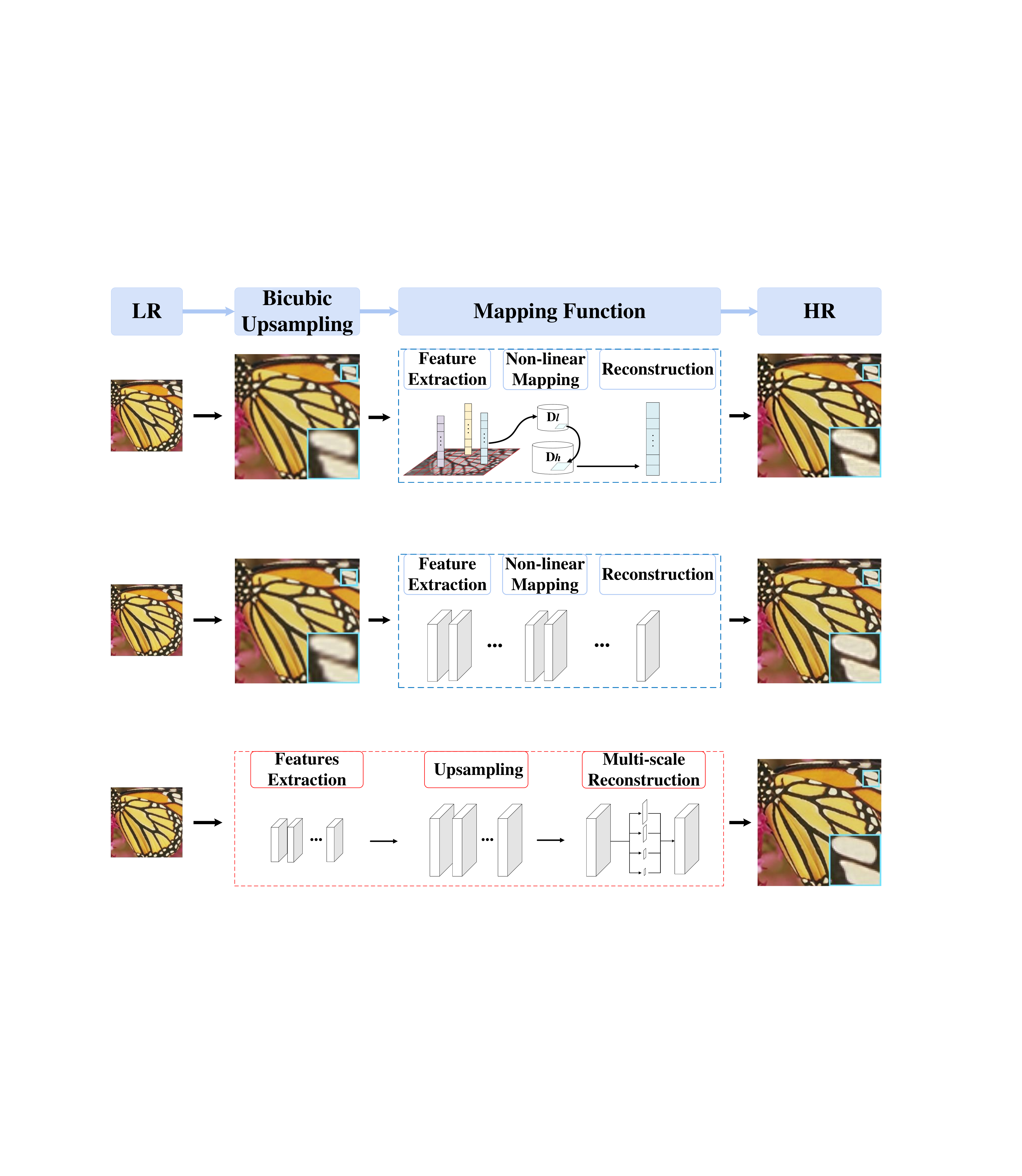}\\
\small{(b)}\\
\includegraphics[width=1.02 \linewidth]{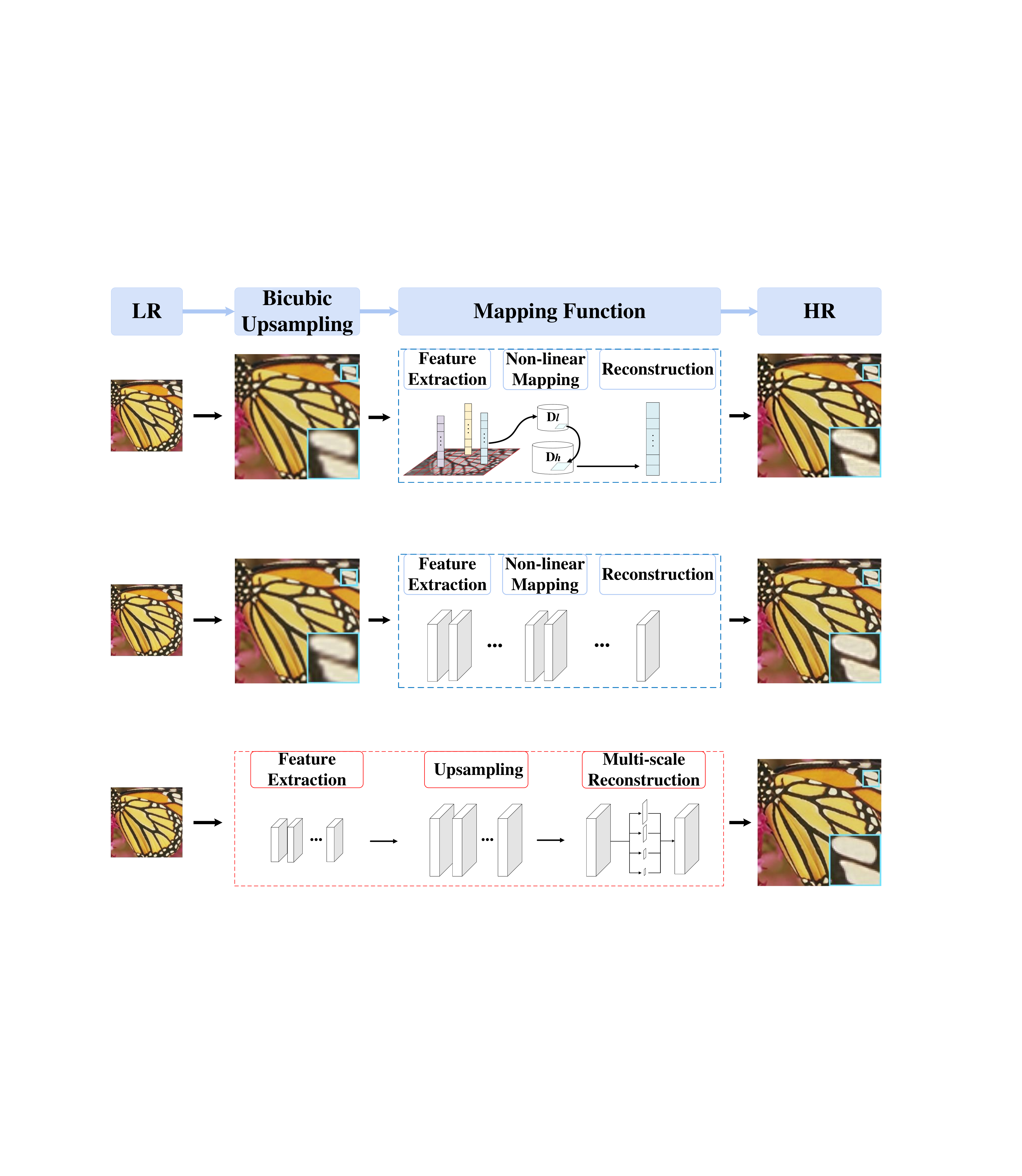}\\
\small{(c)}\\
\end{tabular}

\caption{Overview of learning based SR methods. (a) Prior learning based methods with hand-designed interpolations, features and shallow models. (b) Prior deep learning based methods comprising hand-designed interpolations and automatically learned deep features. (c) The proposed SR method integrating all three steps into a completely end-to-end trainable deep model.  }
\label{motivation}
\vspace{-4mm}
\end{figure}

%
As one of the classical yet challenging problems in image processing, single image super-resolution (SR) aims at restoring the high resolution~(HR) image with abundant high-frequency details from the low resolution (LR) observation.
Given that multiple HR images can be down-sampled into the same LR image, SR as the reverse problem is inherently ill-posed with insufficient knowledge.
Recently, learning-based methods have attracted increasingly more attention from the community and delivered superior performance in image SR.
%
The basic idea is to learn the mapping function from the LR image to the HR image using auxiliary data (Fig.~\ref{motivation} (a)), collected either internally from the input LR image itself (internal-based~\cite{glasner2009super,freedman2011image,yang2013fast}) or externally from abundant natural images (external-based~\cite{freeman2000learning,freeman2002example,yang2010image,zeyde2010single}).
%
%
%
%
A variety of machine learning algorithms, \emph{e.g.}, sparse coding ~\cite{yang2010image,zeyde2010single,yang2012coupled}, anchored neighbor~\cite{chang2004super,yang2013fast,timofte2013anchored,timofte2014a+}, regression trees or forests~\cite{Schulter2015forests,huang2015fast,Salvador2015naive}, have been adopted to learn the mapping function. Though significant progress has been achieved, most of them rely on hand-designed features to characterize LR images, which are not jointly learned with the models. Furthermore, these methods adopt shallow models with limited learning capacity. Both of the two factors may lead to unsatisfactory results and limit further improvements of their performance.
%
%

%
%

More recently, deep neural networks~(DNNs), especially deep convolutional neural networks~(CNNs), have delivered record breaking performance in various computer vision tasks, including image classification, object detection, semantic segmentation, \emph{etc}. Some initial attempts~\cite{dong2014learning,dong2016image,osendorfer2014image,wang2015deep,gu2015convolutional,Kim2016deeply,Kim2016accurate}~have also been made to apply CNNs to image SR.
%
%
%
Most of these methods consist of two decoupled steps (\emph{cf.} Fig.~\ref{motivation} (b)):
 the bicubic interpolation is firstly performed to up-scale the LR image to the size of its HR counterpart;
the CNNs then take the upscaled LR image as input and reconstruct the HR image.
Compared to shallow models, CNNs have much stronger learning capacity, ensuring more accurate prediction of HR images. Furthermore, instead of using hand-designed features, CNNs can automatically learn rich feature hierarchies in a data-driven manner, which are more suitable for the task of image SR.
%

%

	
While impressive performance has been reported, existing deep learning based SR methods have several drawbacks.
Firstly, the two steps of existing methods are individually conducted in a sequential manner and can not be jointly optimized. The stand-alone upsampling step adopts hand-designed techniques (\emph{e.g.}, bicubic interpolation) which are not specifically designed for SR. As a result, they are suboptimal and may even hurt the key information in the original LR images that is crucial for SR. In addition, the original image domain may not be suitable for upsampling, leading to adverse effects on the feature extraction of CNNs.
In comparison, a completely end-to-end CNN that jointly learns the two steps will be more desirable.



Secondly, due to the ill-posed nature of image SR, reconstructing a pixel may depend on either short- or long-range context information. Most prior approaches~\cite{dong2016image,wang2015deep,gu2015convolutional} rely on small image patches to predict the central pixel value, which is less effective for SR with large upscaling factors. To the best of our knowledge, none of existing methods has explored the explicit integration of both short- and long-range context.

Thirdly, it is widely acknowledged that deeper networks with more complex architectures are able to provide more superior performance, which, however, make the training process more challenging. Most prior methods circumvent this issue with relatively shallow CNNs (no more than five layers)~\cite{dong2014learning,dong2016image,wang2015deep,shi2016real}.  The research in \cite{Kim2016accurate} proposes to train very deep networks for SR and good performance has been achieved. Nevertheless, the exploration of training deeper networks for SR is still very limited.

In this paper, we focus our attention on the above three aspects and propose a new image SR method consisting of three major procedures. Feature extraction is firstly performed to map the original LR image into a deep feature space. We then upsample the deep features to the target spatial size with learned filters. Finally, the HR image is reconstructed by considering multi-scale context information of the upsampled deep features. All three procedures are integrated in a single deep CNN model and jointly learned in a data-driven fashion, allowing the image SR to be done in a fully end-to-end manner (\emph{cf.} Fig.~\ref{motivation} (c)). In contrast to prior deep learning based SR approaches, our method automatically learns the upsampling operation in a feature space rather than performing a hand-designed upsampling procedure in the image domain. The joint learning of feature extraction, upsampling, and reconstruction ensures that the feature space is more suitable for upsampling, and the upsampling operation is optimal for the image SR task. In addition, the multi-scale reconstruction combines both short- and long-range context information, allowing more accurate restoration of HR images. We empirically verify that all these proposed techniques can considerably improve performance.

When training the proposed deep CNN for image SR, one critical issue encountered is the exploding/vanishing gradients~\cite{glorot2010understanding}. The shortcut connections proposed in deep residual networks~\cite{he2016} are adopted and can partially address this issue.
As observed in our experiments, although the trained network can capture most high-frequency content in HR images, there exists a discrepancy between the mean magnitudes of the predicted and ground truth HR images. As shown in Fig.~\ref{fig:illuminance}, the overall illumination of the HR image generated by the deep network diverges from that of the ground truth image. The discrepancy constantly varies through the training process and can hardly vanish. In contrast, similar phenomenon is not observed in training relatively shallow networks. The results (Fig.~\ref{fig:illuminance}) produced by a shallow network is more accurate in terms of overall magnitudes. However, limited by its representative capacity, shallow network fails to restore redundant high-frequency details.
%
Motivated by the above observations, we propose to jointly train an ensemble network comprising the proposed deep CNN and a relatively shallow CNN. The shallow CNN can converge more quickly and capture the principle component, \emph{i.e.}, mostly low-frequency content, of the HR images, whereas the deep CNN effectively recovers the high-frequency details. Their combination facilitates faster training and significantly improves performance (\emph{cf.} Fig.~\ref{fig:illuminance}).

\begin{figure}[top]
\centering
\tabcolsep1pt \renewcommand{\arraystretch}{0.9}
\begin{tabular}{cccc}
\includegraphics[width=0.24 \linewidth]{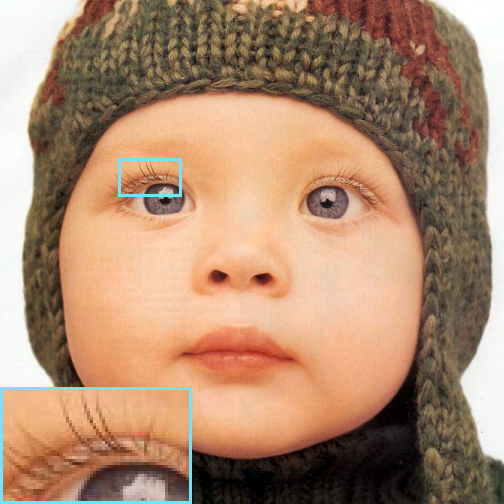}
&\includegraphics[width=0.24 \linewidth]{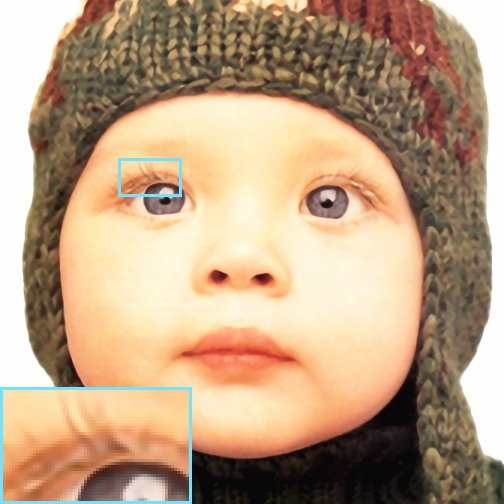}
&\includegraphics[width=0.24 \linewidth]{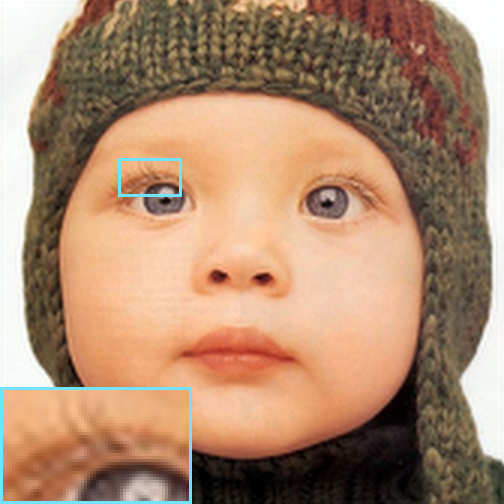}
&\includegraphics[width=0.24 \linewidth]{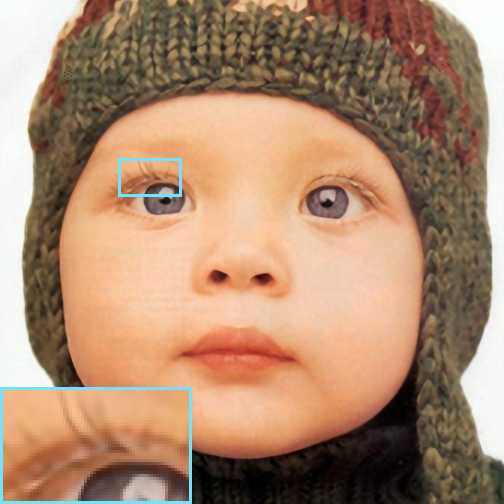}
\\
\includegraphics[width=0.24 \linewidth]{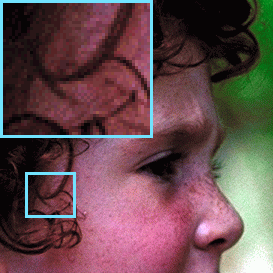}
&\includegraphics[width=0.24 \linewidth]{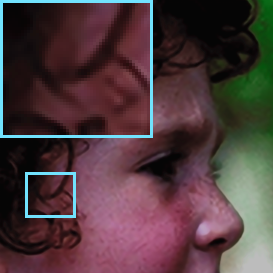}
&\includegraphics[width=0.24 \linewidth]{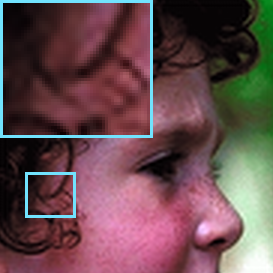}
&\includegraphics[width=0.24 \linewidth]{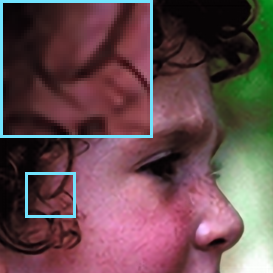}\\
{\small Ground Truth} & {\small Deep} & {\small Shallow} & {\small Ensemble}\\
\end{tabular}

\caption{Super-resolution results of different networks with an upscaling factor $3$. The illumination difference between results of the deep network and the ground truth varies across different images. Results of the shallow network fail to restore high-frequency details. In contrast, the proposed ensemble of deep and shallow networks is able to produce perceptually more accurate results.}
\label{fig:illuminance}
\vspace{-4mm}
\end{figure}

The contributions of this paper are summarized as follows:
\begin{itemize}
\item We propose a fully end-to-end SR method with CNNs, where the upsampling operation is performed in a deep feature space and jointly optimized with the other modules. The HR reconstruction is implemented in a multi-scale manner, combining short- and long-range context.
\item We explore a new deep CNN training approach by jointly training both deep and shallow CNNs as an ensemble, which accelerates training and yields more superior performance. To facilitate interpretation, detailed discussions on our training approach and its relationship to prior residual prediction based SR methods are provided.
\item Extensive evaluations on widely adopted data sets have been conducted to verify the above two contributions. In addition, we perform in-depth analysis on the impact of different network architectures from the perspective of SR. Our findings provide empirical knowledge on CNN architecture design for future image SR research.%
\end{itemize}

\begin{figure*}[top]
\centering
\includegraphics[width=1 \linewidth]{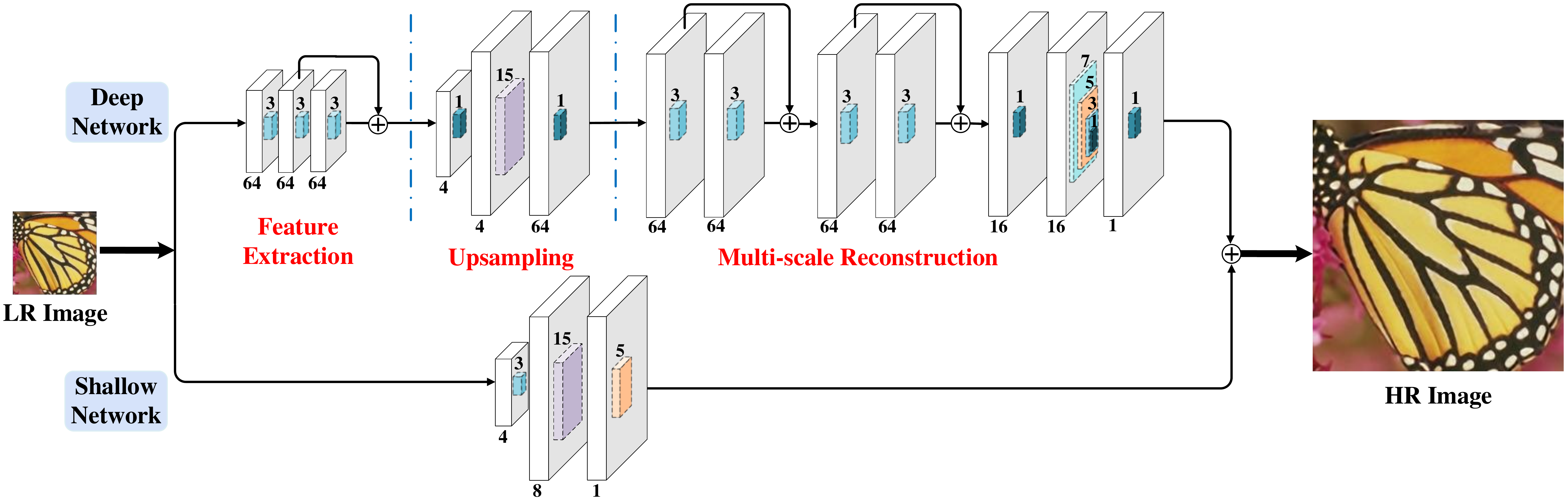}
\caption{Network architecture of the proposed End-to-End Deep and Shallow (EEDS) networks.}
\vspace{-4mm}
\label{fig:pipeline}
\end{figure*}
\section{Related Work}
\label{sec:related work}
%
Image SR can be generally classified into three categories, \emph{i.e.}, interpolation-based~\cite{keys1981cubic,duchon1979lanczos}, reconstruction based~\cite{sun2008image,protter2009generalizing,zhang2012single}, and learning-based methods~\cite{freedman2011image,yang2013fast,huang2015fast}. This paper mainly concentrates on learning-based methods. The basic idea is to formulate image SR as a nonlinear mapping from LR to HR images. The mapping functions can then be approximated by machine learning algorithms in a supervised manner. For instance, Yang~\emph{et al.}~\cite{yang2010image} perform SR by training the LR and HR dictionaries jointly with the constraint that LR patches and its corresponding HR counterparts share the same sparse representation.
In \cite{chang2004super,bevilacqua2012low}, SR is conducted via a neighbor embedding algorithm assuming that LR and HR patches lie on low-dimensional nonlinear manifolds with locally similar geometry. To further improve computational efficiency, Yang and Yang~\cite{yang2013fast}~cluster the LR feature space into numerous subspaces and learn simple mapping functions for each subspace, whereas Timofte~\emph{et al.}~\cite{timofte2013anchored,timofte2014a+}~propose to use a number of linear regressors to locally anchor the neighbors.
%
It is worth noticing that instead of directly mapping LR to HR images, most of the existing approaches~\cite{timofte2013anchored,timofte2014a+} learn regressors to predict the residual between HR and LR by leveraging the fact that LR and HR images are highly correlated, which makes the learning process more tractable.

Recently, deep learning based methods have also been applied to image SR and delivered compelling performance.  In~\cite{dong2014learning}, a CNN is proposed for image SR, which comprises three
convolutional layers corresponding to patch extraction, non-linear mapping and reconstruction, respectively. Later on, Wang~\emph{et al}~\cite{wang2015deep} reformulate traditional sparse coding based method with deep networks, and achieve promising results.
Similar idea has also been explored in a parallel work~\cite{gu2015convolutional}, where convolutional sparse coding is applied to the whole input image rather than individual image patches as in the traditional approaches.

Compared to shallow models, DNNs allow automatically learning feature hierarchies in a data-driven fashion. Their expressive power and strong learning capacity ensure more accurate restoration of abundant high-frequency content. However, training a very deep network for image SR is still a challenging task. Consequently, most prior methods utilize relatively shallow networks.
Inspired by the residual prediction based methods~\cite{zeyde2010single,timofte2013anchored,timofte2014a+}, Kim~\emph{et al.}~\cite{Kim2016accurate} mitigate this issue by training a deep CNN to learn the residual between the HR image and the LR images upscaled with bicubic interpolation. Image SR can then be achieved by combining the upscaled LR image and the predicted residual. In contrast to \cite{Kim2016accurate}, we explore a more principled method by jointly training deep and shallow networks, where the shallow network stabilizes training and the deep network ensures accurate HR reconstruction.

Another drawback of the prior DNN based methods is that the upsampling operation is performed in the image domain and utilizes hand-designed approaches which is decoupled to the DNNs.
This has also been noticed by \cite{osendorfer2014image}, where they propose upsampling in a feature space rather than in the image domain. Nevertheless, they still adopt a manually designed upsampling kernel.
The idea of jointly learning the feature extraction and upsampling filters is developed concurrently and independently in \cite{shi2016real}.
Though bearing a similar spirit, our method significantly differs from \cite{shi2016real} in three aspects: i) \cite{shi2016real} emphasizes more on the efficiency and adopts a relatively shallow network, whereas ours exploits a deeper network for more superior performance; ii) \cite{shi2016real} proposes to increase the resolution only at the very end of the network, while we do so in the latent feature space before reconstruction; and iii) we propose to jointly train deep and shallow networks as an ensemble and reconstruct HR images in a multi-scale manner, while these two important contributions have not been explored by \cite{shi2016real} or other existing methods.
\section{Architecture}
\label{sec:architecture}
%
In this section, we introduce the proposed EEDS (End-to-End Deep and Shallow networks) method for image SR.
Fig.\ref{fig:pipeline} overviews the architecture of the network ensemble comprising a deep and a shallow CNN. The deep CNN consists of 13 trainable layers and can be further divided into three modules: feature extraction, upsampling and multi-scale reconstruction. The complex architecture enables more accurate restoration of detailed HR content, but makes the training more challenging. The shallow CNN with a more simple architecture is easier to converge, which aims at stabilizing the training process of the ensemble. We begin with the description of the deep CNN, and then introduce the architecture of the shallow network.
%
\subsection{Feature Extraction}
\label{sec:feature extraction}
%
%
In order to extract local features of high-frequency content, traditional shallow methods  perform feature extraction by computing the first and second order gradients of the image patch, which is equivalent to filtering the input image with hand-designed, high-pass filters.
Rather than manually designing these filters, deep learning based methods automatically learn these filters from training data. However, most prior methods, no matter using shallow or deep models, extract features from the coarse HR images, which is obtained by upsampling the LR images to the HR size with bicubic interpolation. We argue that the bicubic interpolation is not specifically designed for this purpose, and even damage important LR information, which may play a central role in restoring the HR counterparts. Therefore, the proposed method adopts an alternative strategy and performs feature extraction directly on the original LR images with convolution layers.

Our feature extraction module consists of three convolution layers interleaved by Rectified Linear Unites~(ReLUs) acting as nonlinear mappings. Each convolution layer can be expressed as
%
%
%
%
\begin{equation} \label{eq:conv1}
F_{l}=max(0, W_l \ast F_{l-1}+b_l),
\end{equation}
where $W_l$ and $b_l$ denote the convolution filter and bias of the $l$-th layer, respectively; $F_{l}$ indicates the output feature map of the $l$-th layer, with $F_0$ denoting the original LR images.
All three convolution layers have the same kernel size of $3 \times 3$ and generate feature maps of 64 channels. Zero padding is adopted to preserve the spatial size of the output feature maps.


 %
Inspired by~\cite{he2016}, a shortcut connection with identity mapping is used to add the output feature map of the first layer to the output of the third layer. As justified by \cite{he2016}, the shortcut connections can effectively facilitate gradients flow through multiple layers, thus accelerating deep network training. Similar shortcut structures have also been used in our reconstruction module (Section~\ref{sec:reconstruction}).
%
%

%
\subsection{Upsampling}
\label{sec:upsampling}
Given the extracted features from the original LR images, upsampling operation is performed to increase their spatial span to the target HR size. Instead of using hand-designed interpolation methods, we prefer a learning based upsampling operation, giving rise to an end-to-end trainable system.  To this end, we consider two different strategies widely adopted in CNN for upsampling, \emph{i.e.}, unpooling and deconvolutions. As opposed to pooling layers, the unpooling operation with an upscaling factor $s$ replaces each entry in the input feature map with a $s\times s$ block, where the top left element in the block is set to the value of the input entry and the others to zero. The unpooling operation yields enlarged yet sparse output feature maps. The sparsely activated output values can then be propagated to local neighborhoods by subsequent convolution layers. The deconvolution layer upscales the input feature maps by $s$-fold through reversing the forward and backward propagation of convolution layers with an output stride of $s$. Although unpooling and deconvolution resort to different implementations, they are essentially similar in upscaling feature maps and both are well suited to our task. We adopt the deconvolution layer and achieve promising performance.

The upsampling module connects the feature extraction and reconstruction modules and plays a key role in the proposed SR method. Our experiments empirically show that properly increasing the kernel size of the deconvolution layer can enhance the upsampling quality, leading to an improvement of the final performance. This may be attributed to the fact that a larger deconvolution kernel size allows the upsampling operation to consider a larger input neighborhood and better enforces spatial consistency. Meanwhile, the larger kernel size significantly increases the computational overhead. For both effectiveness and efficiency, the kernel sizes are set to $14\times 14$, $15\times 15$, and $16\times 16$ for upsampling factors of 2, 3, and 4, respectively. Accordingly, zero padding of 6 pixels is conducted on each side of the output feature maps to preserve the spatial size. In addition, two $1 \times 1$ convolutions are conducted before and after the expensive deconvolution layer to further reduce the computational complexity, where the first convolution layer performs dimension reduction by mapping the 64-channel input feature map to the 4-channel output feature map for upsampling, and the last convolution layer then restores the upsampled feature map back to 64 channels. In such a way, the deconvolution operation is performed in a reduced dimension. A ReLU layer is added to the end of the upsampling module to increase non-linearity.

\subsection{Multi-scale Reconstruction}
\label{sec:reconstruction}
Multi-scale inference has been intensively studied in vision problems~\cite{lazebnik2006beyond,he2014spatial,szegedy2015going} and is shown to effectively aggregate local information, allowing more robust and accurate predictions. However, multi-scale inference has been seldom explored in image SR.
Considering that HR image restoration may rely on both short- and long-range context information, we propose to perform HR reconstruction with multi-scale convolutions to explicitly encode multi-context information.

Our HR reconstruction module consists of 7 trainable layers interleaved by ReLU layers for nonlinearity. The first four layers are $3 \times 3$ convolution layers, with each layer taking the previous feature map as input and generating a new feature map of 64 channels. In order to facilitate gradient flow in the training stage, every two convolution layers form a block, where the input is added to the output of the block through a shortcut connection with identity mapping.
The 5-th layer is the dimension reduction layer consists of a $1 \times 1$ convolution, mapping the input feature map of 64 channels to the output 16 channels.
The subsequent multi-scale convolution layer comprises 4 convolution operations of $1\times 1$, $3\times 3$, $5\times 5$, and $7\times 7$ kernel sizes, respectively. All four convolutions are simultaneously conducted on the input feature map and produce four feature maps of 16 channels. The feature maps are then concatenated into a single 64 channel feature map, such that features encoding context information in different scales are fused together. The concatenated feature map is then fed into another $1 \times 1$ convolution layer, which serves as a weighted combination of multi-context feature and reconstruct the final HR images.

\subsection{Combining Deep and Shallow Networks}
\label{sec:combination}
The proposed 13-layer deep network formulates SR as an end-to-end trainable system by directly mapping original LR images to the HR ones. Although shortcut connections have been adopted to alleviate vanishing/exploding gradients, training the proposed deep network is still very challenging. We conjecture the reason may be that the generative nature of the SR problem aims at precisely restoring redundant high-frequency details, as opposed to classification problems, which map different input into the same label space. This challenge becomes even more prominent when our method attempts to directly restore HR details from original LR images rather than those already been upscaled to the target sizes.

In our preliminary experiments, the proposed deep CNN takes longer training time to converge than a relatively shallow CNN. Although the trained deep CNN can more accurately render the high-frequency content of the HR images, the overall magnitudes of the output are inconsistent with those of the ground truth. The illumination errors (Fig.~\ref{fig:illuminance}) in the prediction vary throughout the training process and across different images, and can hardly be corrected by the deep CNN. In comparison, the shallow CNN is able to restore the overall illumination but fails to capture high-frequency details. To combine the best of both worlds, we jointly train an ensemble comprising the proposed deep CNN and another shallow CNN. The shallow CNN facilitates faster convergence and predicts the major component of HR images, while the deep CNN restores high-frequency details and corrects errors of the shallow CNN.



The shallow network consists of three trainable layers corresponding to the three modules of the proposed deep network. The first layer takes the original LR image as input and conducts $3\times 3$ convolutions, producing a feature map of 4 channels. The second layer is a deconvolution layer which upsamples the input feature map to the target spatial size. The final layer reconstruct the HR images from the upsampled feature maps by $5 \times 5$ convolutions.

The deep and shallow networks do not share weights. Both of them independently conduct image SR by taking the same original LR image as input and can be viewed as an ensemble of networks. The final HR image is obtained by
\begin{equation}\label{output}
\hat{Y} = H_{D}(X, \theta_{D}) + H_{S}(X, \theta_{S}),
\end{equation}
where $X$ denotes the input LR image; $H_D(\cdot, \theta_{D})$ and $H_S(\cdot, \theta_{S})$ indicate the HR output of deep and shallow networks parameterized by $\theta_{D}$ and $\theta_{S}$, respectively; $\hat{Y}$ is the final HR image predicted by the ensemble. Detailed discussions on the individual performances and relationships between the deep and shallow networks are provided in Section~\ref{sec:analysis}.

\ignore{
\begin{figure}[top]
\centering
\tabcolsep1pt \renewcommand{\arraystretch}{0.9}
\begin{tabular}{cccc}
\includegraphics[width=0.24 \linewidth]{fig/Motivation_2/baby/baby_x3_GT.png}
&\includegraphics[width=0.24 \linewidth]{fig/Motivation_2/baby/baby_x3_deep.png}
&\includegraphics[width=0.24 \linewidth]{fig/Motivation_2/baby/baby_x3_shallow.png}
&\includegraphics[width=0.24 \linewidth]{fig/Motivation_2/baby/baby_x3_joint.png}
\\
\includegraphics[width=0.24 \linewidth]{fig/Motivation_2/head/head_x3_GT.png}
&\includegraphics[width=0.24 \linewidth]{fig/Motivation_2/head/head_x3_deep.png}
&\includegraphics[width=0.24 \linewidth]{fig/Motivation_2/head/head_x3_shallow.png}
&\includegraphics[width=0.24 \linewidth]{fig/Motivation_2/head/head_x3_joint.png}\\
{\small (a)} & {\small(b)} & {\small(c)} & {\small(d)}\\
\end{tabular}

\caption{Super-resolution results of different networks with an upscaling factor $3$. (a) Ground truth, (b) the deep network, (c) the shallow network, (d) the combination of deep and shallow networks. Note the illumination differences between the outputs of the deep network and the ground truth. }
\label{fig:illuminance}
\end{figure}
}
\subsection{Training}
\label{sec:training}
Given $N$ training image pairs $\{X_i,Y_i\}_{i=1}^N$, the proposed deep and shallow networks are jointly learned by minimizing the Euclidean loss between the predicted HR image $\hat{Y}$ and the ground truth $Y$:
\begin{equation}
\theta_{D}^*, \theta_{S}^* = \arg \min_{\theta_{D},\theta_{S}} \frac{1}{2N}\sum_{i=1}^{N} \| \hat{Y}_i - Y_i\|_2^2 + \eta R(\theta_{D}, \theta_{S}),
\end{equation}
where $\hat{Y}$ is the predicted HR image computed as Eq.~\eqref{output}, and $R(\theta_{D}, \theta_{S})$ denotes the weight decay imposed on network parameters with a small trade off $\eta$.
%
%
%
%

The optimization is conducted by the mini-batch stochastic gradient descent method with a batch size of $256$, momentum of $0.9$, and weight decay of $0.005$.
%
All the filters in convolution layers are randomly initialized from a zero-mean Gaussian distribution with standard deviation $0.01$. The filters in deconvolution layers are initialized from bilinear interpolation kernels. 
The learning rate is initially set to $1e-4$ and decreased by a factor of 0.1 when the validation loss is stabilized.
%

%

\section{Experiments}
\label{sec:experiment}
%
\subsection{Setup}
\label{sec:setup}
For fair comparisons with existing methods, we use the same training sets, test sets and protoclos which are widely-adopted~\cite{dong2014learning,wang2015deep,gu2015convolutional}.
Specifically, our model is trained using 91 images proposed in \cite{yang2010image}. To avoid over-fitting and further improve accuracy, data augmentation techniques including rotation and flipping are performed, yielding a training set of 728 images and a validation set of 200 images.
For each upscaling factor (\emph{i.e.}, 2, 3 or 4), $96\times 96$ patches are randomly cropped from training images as the ground truth HR examples, which are then downsampled using the bicubic interpolation to generate the corresponding LR training samples. The proposed model is trained using the Caffe package~\cite{jia2014caffe} on a workstation with a Intel $3.6$ GHz CPU and a GTX980 GPU. The training takes approximately 95 epochs to converge. The source code and trained models to reproduce the experimental results will be released upon acceptance of the submission.

We evaluate the performance of upscaling factors 2, 3 and 4 on three public datasets:
Set5~\cite{bevilacqua2012low}, Set14~\cite{zeyde2010single} and BSD100~\cite{martin2001database}, which contain 5, 14, and 100 images, respectively. We utilize PSNR and SSIM~\cite{wang2004image} metrics for quantitative evaluation, which are widely used in the image SR literature. At inference, our model takes the original LR image of arbitrary size as input and directly reconstructs the corresponding HR image.

Since humans are more sensitive to changes of luminance than color, we follow most existing methods and only super-resolve the luminance channel in YCbCr color space.
For the purpose of displaying, the other two chrominance channels are simply upsampled by bicubic interpolation.
%

%
\subsection{Comparison with state-of-the-arts}
\label{sec:comparison}
%
We compare the proposed EEDS model with state-of-the-art methods for three upscaling factors~(2, 3, 4) both qualitatively and quantitatively.
The compared methods include the traditional bicubic interpolation and nine popular learning based methods: the scale-up using sparse representation~(SUSR)~proposed by Zeyde~\emph{et al.}~\cite{zeyde2010single}, adjusted anchored neighborhood regression method (A$+$)~\cite{timofte2014a+}, two forests based methods, \emph{i.e.}, ARFL~\cite{Schulter2015forests} and NBSRF~\cite{Salvador2015naive}, five deep learning based methods SRCNN~\cite{dong2014learning}, SRCNN-L~\cite{dong2016image}, CSC~\cite{gu2015convolutional}, CSCN~\cite{wang2015deep}, and ESPCN~\cite{shi2016real}.
%
%
The results of the compared methods are either obtained using the publicly available codes or provided by the authors.

Table~\ref{tab:compare} summaries the quantitative performance of compared methods measured by average PSNR and SSIM \footnote{The performance of ESPCN(ImageNet) and SRCN-L are also included in Table~\ref{tab:compare} for reference. Since both methods are trained on the ImageNet data set, they are not counted for performance ranking.}. The proposed EEDS method consistently outperforms the other methods across three test sets for all upscaling factors. As demonstrated in the last line of Table~\ref{tab:compare}, our method improves the performance over the second best method (CSCN) by a considerable margin in terms of both PSNR and SSIM. It should be noted that the CSCN method adopts a cascaded strategy to conduct SR for the upscaling factor of 3 and 4 (\emph{i.e.}, by super-resolving the LR image twice with a factor of 2), which is shown to improve the final performance. The ESPCN (ImageNet) method is trained using 50,000 images from ImageNet data set\cite{imagenet09}. Further performance improvements of our method can also be expected when using the cascaded strategy or with more training images.

Fig.~\ref{fig:comic} to Fig.~\ref{fig:butterfly} illustrate some sampled results generated by the compared methods. The HR images restored by the proposed EEDS method are perceptually more plausible with relatively sharp edges and little artifacts.
%

%
%


\begin{table*}[htbp]
\linespread{1}
\renewcommand\arraystretch{1.3}
\renewcommand{\multirowsetup}{\centering}
\caption{Average PSNR(SSIM) comparison on three test datasets among different methods. \R{Red} and {\color{blue}{blue}} colors indicate the best and the second best performance.}
\label{tab:compare}
\centering
\begin{tabular}{|c||c|c|c||c|c|c||c|c|c|}
\hline
Dataset & \multicolumn{3}{c||}{Set5} & \multicolumn{3}{c||}{Set14} & \multicolumn{3}{c|}{BSD100} \\
\hline
Scale   & x2 & x3 & x4   & x2 & x3  & x4   & x2 & x3  & x4  \\                                                    \hline \hline
Bicubic  & \begin{tabular}[c]{@{}c@{}}33.66 \\ (0.9299)\end{tabular} & \begin{tabular}[c]{@{}c@{}}30.39\\ (0.8682)\end{tabular} & \begin{tabular}[c]{@{}c@{}}28.42\\ (0.8104)\end{tabular}  & \begin{tabular}[c]{@{}c@{}}30.24\\ (0.8687)\end{tabular} & \begin{tabular}[c]{@{}c@{}}27.55\\ (0.7736)\end{tabular} & \begin{tabular}[c]{@{}c@{}}26.00\\ (0.7019)\end{tabular}  & \begin{tabular}[c]{@{}c@{}}29.56\\ (0.8431)\end{tabular} & \begin{tabular}[c]{@{}c@{}}27.21\\ (0.7385)\end{tabular} & \begin{tabular}[c]{@{}c@{}}25.96\\ (0.6675)\end{tabular} \\
\hline
SUSR~\cite{zeyde2010single}  & \begin{tabular}[c]{@{}c@{}}35.78\\ (0.9493)\end{tabular} & \begin{tabular}[c]{@{}c@{}}31.90\\ (0.8968)\end{tabular} & \begin{tabular}[c]{@{}c@{}}29.69\\ (0.8428)\end{tabular}  & \begin{tabular}[c]{@{}c@{}}31.81\\
(0.8988)\end{tabular} & \begin{tabular}[c]{@{}c@{}}28.67\\ (0.8075)\end{tabular} & \begin{tabular}[c]{@{}c@{}}26.88\\ (0.7342)\end{tabular}  & \begin{tabular}[c]{@{}c@{}}30.40\\
(0.8682 )\end{tabular} & \begin{tabular}[c]{@{}c@{}}27.15\\ (0.7695)\end{tabular} & \begin{tabular}[c]{@{}c@{}}25.92\\ (0.6968)\end{tabular} \\
\hline
A$+$~\cite{timofte2014a+}  & \begin{tabular}[c]{@{}c@{}}36.55\\ (0.9544)\end{tabular}  & \begin{tabular}[c]{@{}c@{}}32.59\\ (0.9088)\end{tabular} & \begin{tabular}[c]{@{}c@{}}30.29\\ (0.8603)\end{tabular}  & \begin{tabular}[c]{@{}c@{}}32.28\\ (0.9056)\end{tabular} & \begin{tabular}[c]{@{}c@{}}29.13\\ (0.8188)\end{tabular} & \begin{tabular}[c]{@{}c@{}}27.32\\ (0.7491)\end{tabular}  & \begin{tabular}[c]{@{}c@{}}30.78\\ (0.8773)\end{tabular} & \begin{tabular}[c]{@{}c@{}}28.18\\ (0.7808)\end{tabular} & \begin{tabular}[c]{@{}c@{}}26.77\\ (0.7085)\end{tabular} \\
\hline
ARFL~\cite{Schulter2015forests}   & \begin{tabular}[c]{@{}c@{}}36.71\\  (0.9548)\end{tabular} & \begin{tabular}[c]{@{}c@{}}32.57\\ (0.9077)\end{tabular} & \begin{tabular}[c]{@{}c@{}}30.21\\ (0.8565)\end{tabular}  & \begin{tabular}[c]{@{}c@{}}32.36\\ (0.9059)\end{tabular} & \begin{tabular}[c]{@{}c@{}}29.12\\ (0.8181)\end{tabular} & \begin{tabular}[c]{@{}c@{}}27.31\\ (0.7465)\end{tabular}  & \begin{tabular}[c]{@{}c@{}}31.26\\ (0.8864)\end{tabular} & \begin{tabular}[c]{@{}c@{}}28.28\\ (0.7825)\end{tabular} & \begin{tabular}[c]{@{}c@{}}26.79\\ (0.7066)\end{tabular} \\
\hline
NBSRF~\cite{Salvador2015naive}  & \begin{tabular}[c]{@{}c@{}}36.76\\  (0.9552)\end{tabular} & \begin{tabular}[c]{@{}c@{}}32.75\\ (0.9104)\end{tabular} & \begin{tabular}[c]{@{}c@{}}30.44\\ (0.8632)\end{tabular}  & \begin{tabular}[c]{@{}c@{}}32.45\\ (0.9071)\end{tabular} & \begin{tabular}[c]{@{}c@{}}29.25\\ (0.8212)\end{tabular} & \begin{tabular}[c]{@{}c@{}}27.41\\ (0.7511)\end{tabular}  & \begin{tabular}[c]{@{}c@{}}31.30\\ (0.8876)\end{tabular} & \begin{tabular}[c]{@{}c@{}}28.36\\ (0.7856)\end{tabular} & \begin{tabular}[c]{@{}c@{}}26.88\\ (0.7110)\end{tabular} \\
\hline
SRCNN~\cite{dong2014learning}  & \begin{tabular}[c]{@{}c@{}}36.34\\ (0.9521)\end{tabular}  & \begin{tabular}[c]{@{}c@{}}32.39\\ (0.9033)\end{tabular} & \begin{tabular}[c]{@{}c@{}}30.09\\ (0.8530)\end{tabular}  & \begin{tabular}[c]{@{}c@{}}32.18\\ (0.9039)\end{tabular} & \begin{tabular}[c]{@{}c@{}}29.00\\ (0.8145)\end{tabular} & \begin{tabular}[c]{@{}c@{}}27.20\\ (0.7413)\end{tabular}  & \begin{tabular}[c]{@{}c@{}}31.11\\ (0.8835)\end{tabular} & \begin{tabular}[c]{@{}c@{}}28.20\\ (0.7794)\end{tabular} & \begin{tabular}[c]{@{}c@{}}26.70\\ (0.7018)\end{tabular} \\
\hline
SRCNN-L~\cite{dong2016image}  & \begin{tabular}[c]{@{}c@{}}36.66\\ (0.9542)\end{tabular}  & \begin{tabular}[c]{@{}c@{}}32.75\\ (0.9090)\end{tabular} & \begin{tabular}[c]{@{}c@{}}30.49\\ (0.8628)\end{tabular}  & \begin{tabular}[c]{@{}c@{}}32.45\\ (0.9067)\end{tabular} & \begin{tabular}[c]{@{}c@{}}29.30\\ (0.8215)\end{tabular} & \begin{tabular}[c]{@{}c@{}}27.50\\ (0.7513)\end{tabular}  & \begin{tabular}[c]{@{}c@{}}31.36\\ (0.8879)\end{tabular} & \begin{tabular}[c]{@{}c@{}}28.41\\ (0.7863)\end{tabular} & \begin{tabular}[c]{@{}c@{}}26.90\\ (0.7103)\end{tabular} \\
\hline
CSC~\cite{gu2015convolutional}   & \begin{tabular}[c]{@{}c@{}}36.62\\ (0.9548)\end{tabular}  & \begin{tabular}[c]{@{}c@{}}32.66\\ (0.9098)\end{tabular} & \begin{tabular}[c]{@{}c@{}}30.36\\ (0.8607)\end{tabular}  & \begin{tabular}[c]{@{}c@{}}32.31 \\ (0.9070)\end{tabular}      & \begin{tabular}[c]{@{}c@{}}29.16\\ (0.8209)\end{tabular} & \begin{tabular}[c]{@{}c@{}}27.30\\ (0.7499)\end{tabular}  & \begin{tabular}[c]{@{}c@{}}31.27\\(0.8876)\end{tabular}       & \begin{tabular}[c]{@{}c@{}}28.31\\ (0.7853)\end{tabular} & \begin{tabular}[c]{@{}c@{}}26.83\\ (0.7101)\end{tabular} \\
\hline
ESPCN~\cite{shi2016real} &  $-$  & \begin{tabular}[c]{@{}c@{}}32.55\\ ($-$)\end{tabular} &$-$ &
 $-$ & \begin{tabular}[c]{@{}c@{}}29.08\\ ($-$)\end{tabular} & $-$ & $-$ & $-$ & $-$ \\
\hline
ESPCN (ImageNet)~\cite{shi2016real} & $-$ & \begin{tabular}[c]{@{}c@{}}33.13\\ ($-$ )\end{tabular} & \begin{tabular}[c]{@{}c@{}}30.90\\ ($-$)\end{tabular}  & $-$  & \begin{tabular}[c]{@{}c@{}}29.49\\ ($-$)\end{tabular} & \begin{tabular}[c]{@{}c@{}}27.73\\ ($-$)\end{tabular}  & $-$ & $-$ & $-$ \\
\hline
CSCN~\cite{wang2015deep}  & \begin{tabular}[c]{@{}c@{}}{\color{blue}{36.93}}\\ ({\color{blue}{0.9552}})\end{tabular}  & \begin{tabular}[c]{@{}c@{}}{\color{blue}{33.10}}\\ ({\color{blue}{0.9144}})\end{tabular} & \begin{tabular}[c]{@{}c@{}}{\color{blue}{30.86}}\\ ({\color{blue}{0.8732}})\end{tabular}  & \begin{tabular}[c]{@{}c@{}}{\color{blue}{32.56}}\\ ({\color{blue}{0.9074}})\end{tabular} & \begin{tabular}[c]{@{}c@{}}{\color{blue}{29.41}}\\ ({\color{blue}{0.8238}})\end{tabular} & \begin{tabular}[c]{@{}c@{}}{\color{blue}{27.64}}\\ ({\color{blue}{0.7578}})\end{tabular}  & \begin{tabular}[c]{@{}c@{}}{\color{blue}{31.40}}\\ ({\color{blue}{0.8884}})\end{tabular} & \begin{tabular}[c]{@{}c@{}}{\color{blue}{28.50}}\\ ({\color{blue}{0.7885}})\end{tabular} & \begin{tabular}[c]{@{}c@{}}{\color{blue}{27.03}}\\ ({\color{blue}{0.7161}})\end{tabular} \\
\hline
EEDS& \begin{tabular}[c]{@{}c@{}}{\color{red}{37.29}}\\ ({\color{red}{0.9579}})\end{tabular} & \begin{tabular}[c]{@{}c@{}}{\color{red}{33.47}}\\ ({\color{red}{0.9191}})\end{tabular} & \begin{tabular}[c]{@{}c@{}}{\color{red}{31.14}}\\ ({\color{red}{0.8783}})\end{tabular} &  \begin{tabular}[c]{@{}c@{}}{\color{red}{32.81}}\\ ({\color{red}{0.9105}})\end{tabular} & \begin{tabular}[c]{@{}c@{}}{\color{red}{29.60}}\\ ({\color{red}{0.8284}})\end{tabular} & \begin{tabular}[c]{@{}c@{}}{\color{red}{27.82}}\\ ({\color{red}{0.7626}})\end{tabular} &  \begin{tabular}[c]{@{}c@{}}{\color{red}{31.64}}\\ ({\color{red}{0.8928}})\end{tabular} & \begin{tabular}[c]{@{}c@{}}{\color{red}{28.64}}\\ ({\color{red}{0.7925}})\end{tabular} & \begin{tabular}[c]{@{}c@{}}{\color{red}{27.11}}\\ ({\color{red}{0.7200}})\end{tabular} \\
\hline
\hline
Our Gain &   \begin{tabular}[c]{@{}c@{}}0.36\\ (0.0027)\end{tabular}  & \begin{tabular}[c]{@{}c@{}}0.37\\ (0.0047)\end{tabular} & \begin{tabular}[c]{@{}c@{}}0.28\\ (0.0051)\end{tabular} &   \begin{tabular}[c]{@{}c@{}}0.25 \\ (0.0031)\end{tabular}      & \begin{tabular}[c]{@{}c@{}}0.19\\ (0.0046)\end{tabular} & \begin{tabular}[c]{@{}c@{}}0.18\\ (0.0048)\end{tabular} &  \begin{tabular}[c]{@{}c@{}}0.24\\(0.0044)\end{tabular}       & \begin{tabular}[c]{@{}c@{}}0.14\\ (0.0040)\end{tabular} & \begin{tabular}[c]{@{}c@{}}0.08\\ (0.0039)\end{tabular} \\
\hline
\end{tabular}
\end{table*}

\begin{figure*}[htbp]
\centering
\renewcommand\arraystretch{0.9}
\begin{tabular}{@{}c@{}c@{}c@{}c@{}c}
\includegraphics[width=0.2\linewidth]{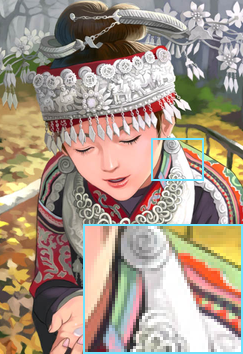} \ &
\includegraphics[width=0.2\linewidth]{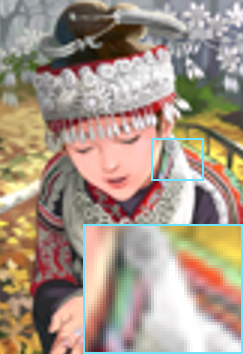} \ &
\includegraphics[width=0.2\linewidth]{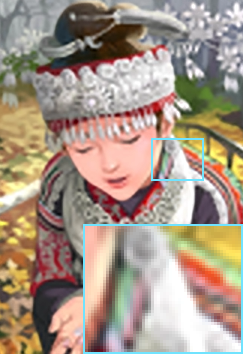} \ &
\includegraphics[width=0.2\linewidth]{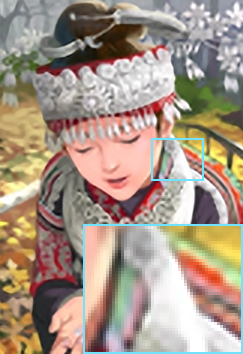} \ &
\includegraphics[width=0.2\linewidth]{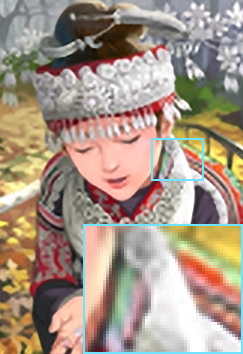} \\
{\small(a) Ground truth} & {\small(b) Bicubic} & {\small(c) SUSR~\cite{zeyde2010single}} &{\small(d) A+~\cite{timofte2014a+}} & {\small(e) ASRF~\cite{Schulter2015forests}} \\
\includegraphics[width=0.2\linewidth]{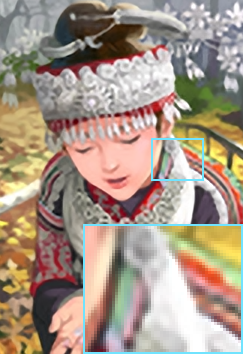} \ &
\includegraphics[width=0.2\linewidth]{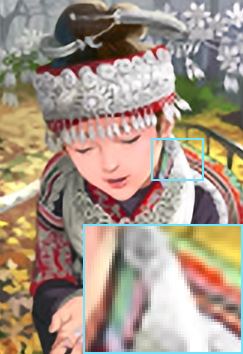} \ &
\includegraphics[width=0.2\linewidth]{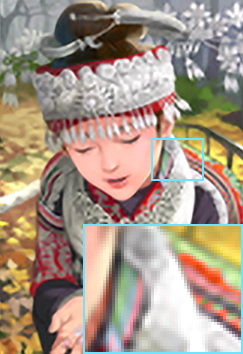} \ &
\includegraphics[width=0.2\linewidth]{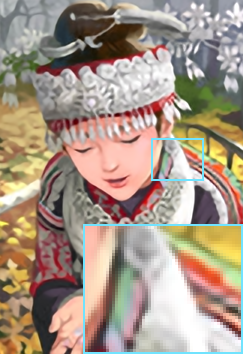} \ &
\includegraphics[width=0.2\linewidth]{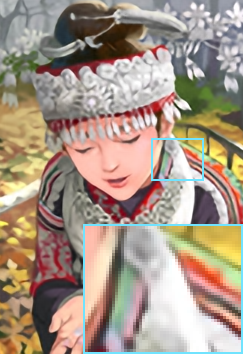}  \\
{\small(f) NBSRF~\cite{Salvador2015naive}} & {\small(g) SRCNN-L~\cite{dong2016image}} & {\small(h) CSC~\cite{gu2015convolutional}} &{\small(i) CSCN~\cite{wang2015deep}} & {\small (j) EEDS} \\
\end{tabular}
\vspace{-2mm}
 {\caption{The “comic” image from Set14 with an upscaling factor 3.}\label{fig:comic}}
\end{figure*}

\begin{figure*}[htbp]
\centering
\renewcommand\arraystretch{0.9}
\begin{tabular}{@{}c@{}c@{}c@{}c@{}c}
\includegraphics[width=0.2\linewidth]{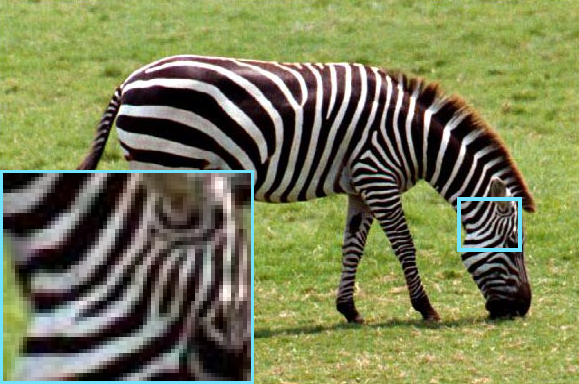} \ &
\includegraphics[width=0.2\linewidth]{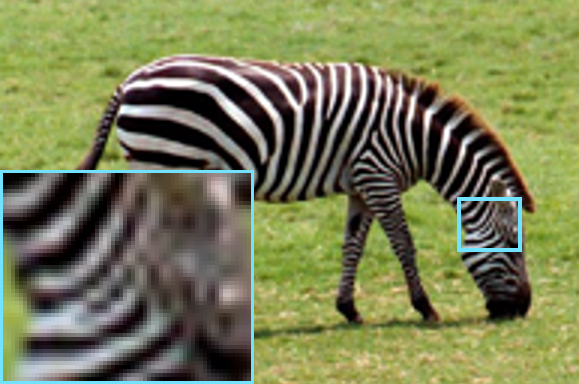} \ &
\includegraphics[width=0.2\linewidth]{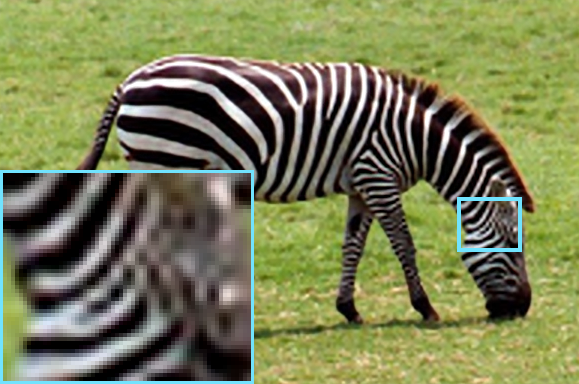} \ &
\includegraphics[width=0.2\linewidth]{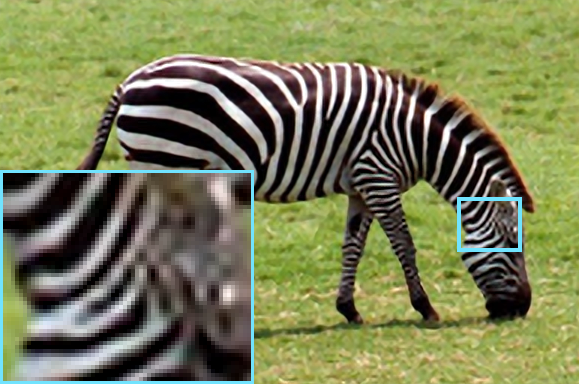} \ &
\includegraphics[width=0.2\linewidth]{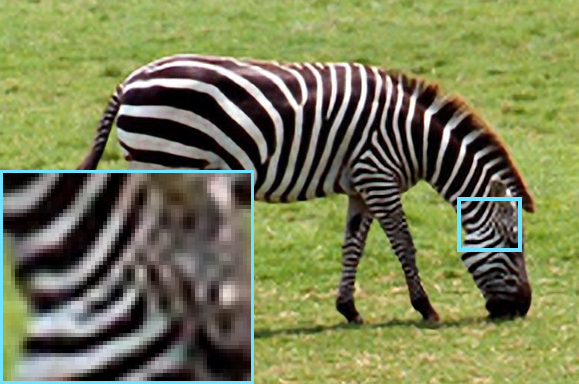} \\
{\small(a) Ground truth} & {\small(b) Bicubic} & {\small(c) SUSR~\cite{zeyde2010single}} &{\small(d) A+~\cite{timofte2014a+}} & {\small(e) ASRF~\cite{Schulter2015forests}} \\
\includegraphics[width=0.2\linewidth]{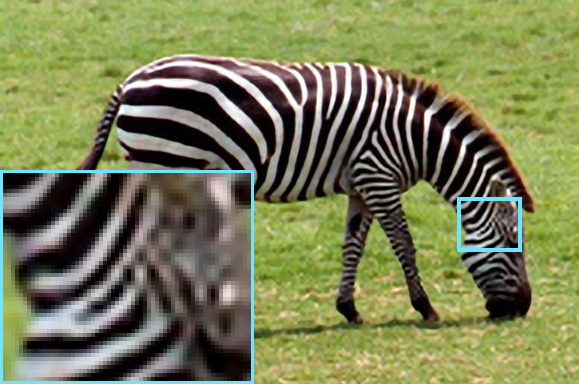} \ &
\includegraphics[width=0.2\linewidth]{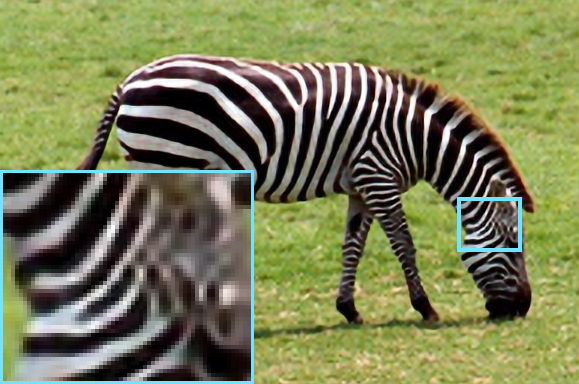} \ &
\includegraphics[width=0.2\linewidth]{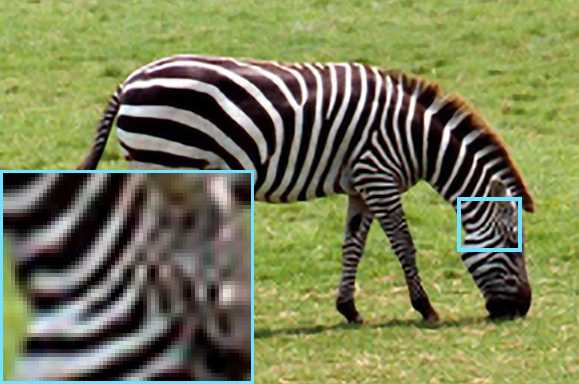} \ &
\includegraphics[width=0.2\linewidth]{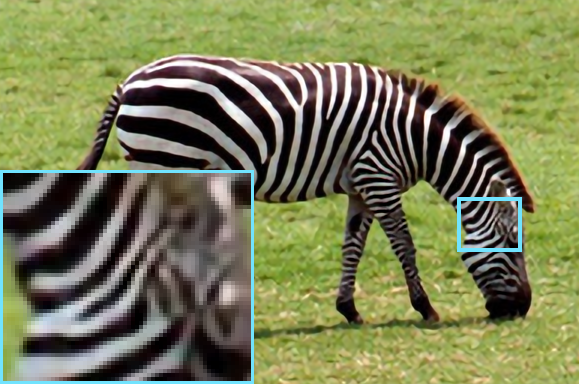} \ &
\includegraphics[width=0.2\linewidth]{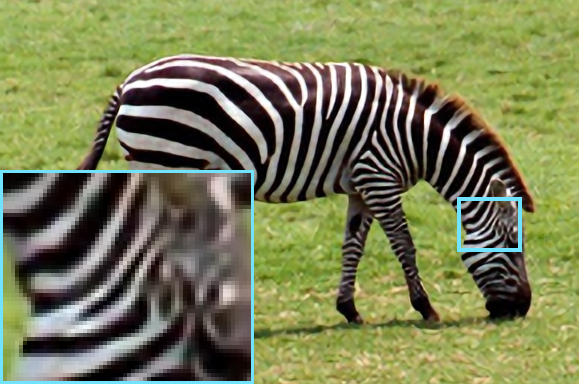} \ \\
{\small(f) NBSRF~\cite{Salvador2015naive}} & {\small(g) SRCNN-L~\cite{dong2016image}} & {\small(h) CSC~\cite{gu2015convolutional}} &{\small(i) CSCN~\cite{wang2015deep}} & {\small (j) EEDS} \\
\end{tabular}
\vspace{-2mm}
 {\caption{The “barbara” image from Set14 with an upscaling factor 3.}

 \label{fig:barbara}}
\end{figure*}

\begin{figure*}[htbp]
\centering
\renewcommand\arraystretch{0.9}
\begin{tabular}{@{}c@{}c@{}c@{}c@{}c}
\includegraphics[width=0.2\linewidth]{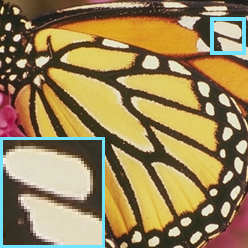} \ &
\includegraphics[width=0.2\linewidth]{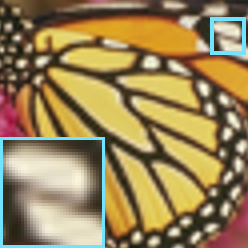} \ &
\includegraphics[width=0.2\linewidth]{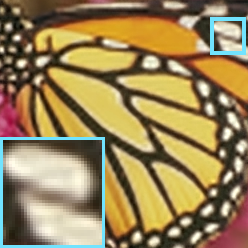} \ &
\includegraphics[width=0.2\linewidth]{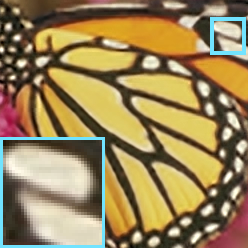} \ &
\includegraphics[width=0.2\linewidth]{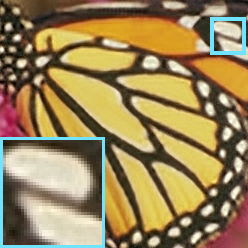}\ \\
{\small(a) Ground truth} & {\small(b) Bicubic} & {\small(c) SUSR~\cite{zeyde2010single}} &{\small(d) A+~\cite{timofte2014a+}} & {\small(e) ASRF~\cite{Schulter2015forests}} \\
\includegraphics[width=0.2\linewidth]{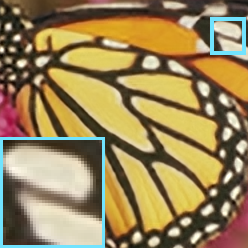} \ &
\includegraphics[width=0.2\linewidth]{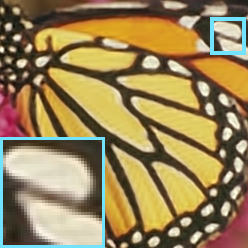} \ &
\includegraphics[width=0.2\linewidth]{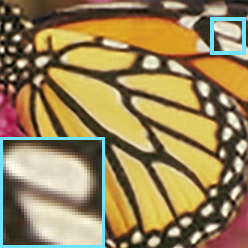} \ &
\includegraphics[width=0.2\linewidth]{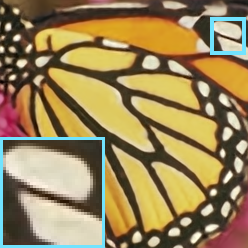} \ &
\includegraphics[width=0.2\linewidth]{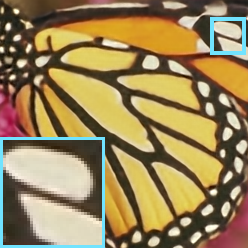} \ \\
{\small(f) NBSRF~\cite{Salvador2015naive}} & {\small(g) SRCNN-L~\cite{dong2016image}} & {\small(h) CSC~\cite{gu2015convolutional}} &{\small(i) CSCN~\cite{wang2015deep}} & {\small (j) EEDS} \\
\end{tabular}
\vspace{-2mm}
{\caption{The “butterfly” image from Set5 with an upscaling factor 4.}\label{fig:butterfly}}
\end{figure*}

\subsection{Architecture Analysis}
\label{sec:analysis}
To gain further insights of our contributions, we conduct additional evaluations on different variants of the proposed EEDS method. Unless stated otherwise, we strictly follow the implementation settings in Section~\ref{sec:setup} to train all the methods. Since similar phenomena are observed for different upscaling factors, we only report the results for the upscaling factor of 3.



\begin{figure}[t]
\centering
\includegraphics[width=0.8\linewidth]{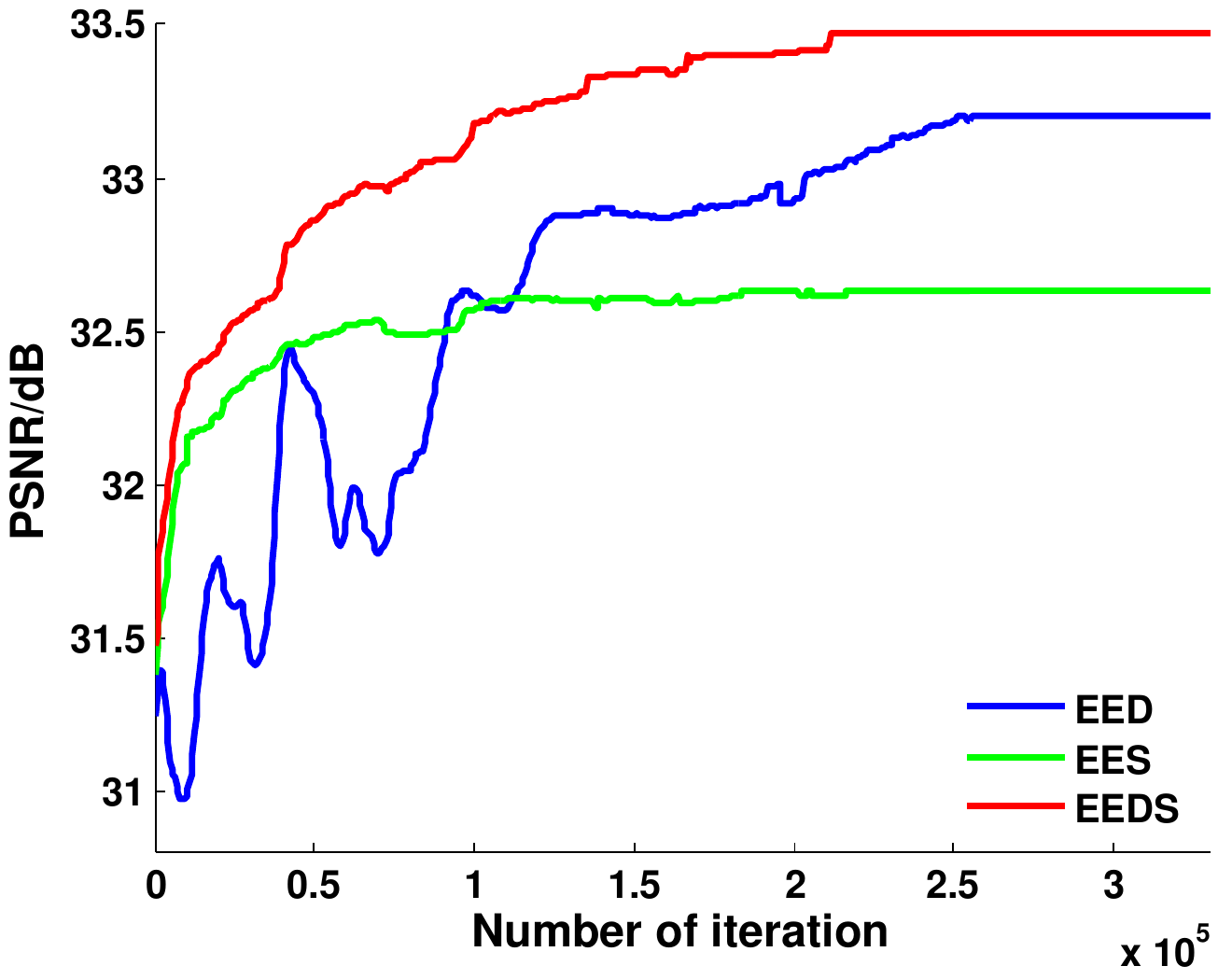}
\caption{Convergence plots of different CNN architectures on the Set5 data set with an upsampling factor 3.}
\label{curve:deep-shallow}
\end{figure}
{\flushleft \textbf{Deep Network vs Shallow Network.} Our method jointly trains a deep and a shallow network as an ensemble. To investigate the impact of the two networks on the final performance, we split the two networks and obtain two variants of the proposed EEDS model, namely, the EED (end-to-end deep network) and EES (end-to-end shallow network), respectively.
Fig.~\ref{curve:deep-shallow} depicts the convergence plots of all three models on the Set5 data set. The EES method with a shallow network takes less time to converge. However, limited by its capacity, the final performance of EES is relatively low. In contrast, the EED is more difficult to train. Upon convergence, the EED method achieves higher PSNR than EES. However, the performance of EED is still not stable. The deep architecture of EED is sufficiently complex to restore high-frequency details. However, the generated HR images suffer from illumination changes compared to the ground truth.  This may be attributed to the fact that directly mapping LR images to HR is a very complex task and EED may converge to some local minimum.

The proposed EEDS method mitigates this issue by combining deep and shallow networks as an ensemble. At joint training, the shallow network still converges much faster and dominates the performance at the very beginning (Fig.~\ref{curve:deep-shallow}). After the shallow network has already captured the major components of the HR images, the difficulty of direct SR has been significantly lowered. The deep network then starts to focus on the high-frequency details and learns to correct the errors made by the shallow network. As shown in Fig.~\ref{curve:deep-shallow}, the EEDS method is much faster to converge than EED and achieves the best performance among all three methods. Upon convergence, the prediction made by the shallow network of EEDS restores most content with blur and artifacts (Fig.~\ref{fig:flow-result} (b)), whereas the output of the deep network of EEDS mostly contains high-frequency content (Fig.~\ref{fig:flow-result} (c)). By combining the two outputs, the EEDS method achieves more accurate results (Fig.~\ref{fig:flow-result} (d)).

\begin{figure}[t]
\centering
\tabcolsep1pt \renewcommand{\arraystretch}{0.9}
\begin{tabular}{cccc}
\includegraphics[width=0.24 \linewidth]{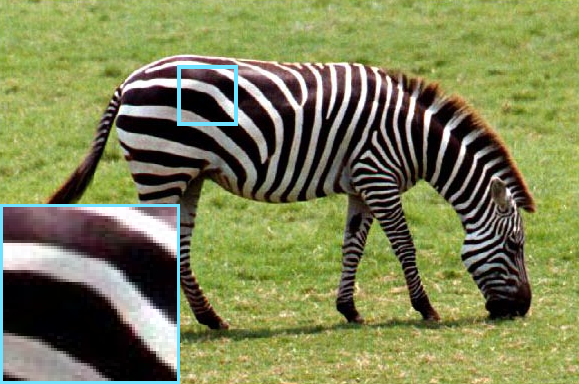}
&\includegraphics[width=0.24 \linewidth]{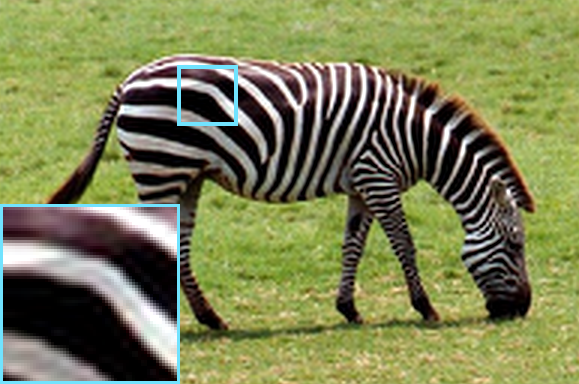}
&\includegraphics[width=0.24 \linewidth]{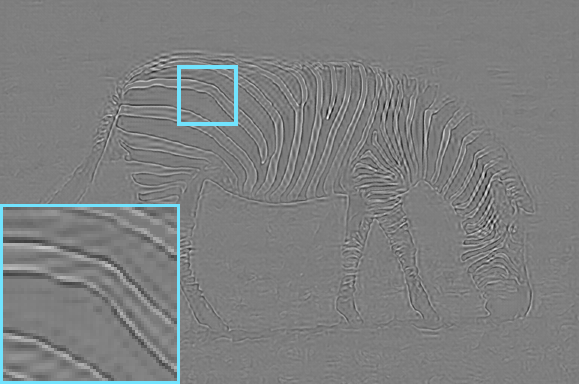}
&\includegraphics[width=0.24 \linewidth]{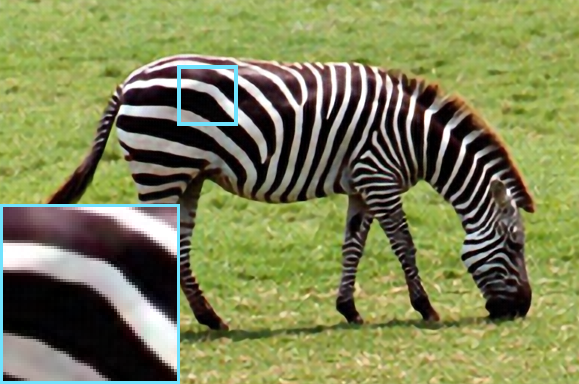}
\\
\includegraphics[width=0.24 \linewidth]{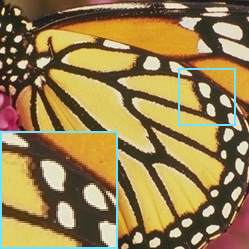}
&\includegraphics[width=0.24 \linewidth]{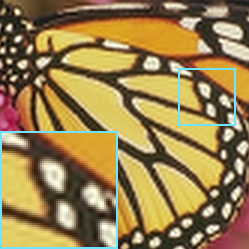}
&\includegraphics[width=0.24 \linewidth]{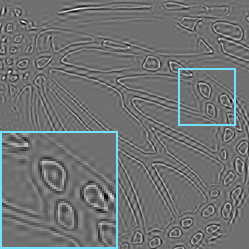}
&\includegraphics[width=0.24 \linewidth]{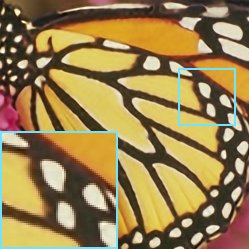}
\\
{\small(a) } & {\small(b)} & {\small(c)} & {\small(d)}\\
\end{tabular}
\caption{Output of the proposed EEDS model and its subnetworks with an upscaling factor 3. (a) Ground truth, (b) output of the shallow network of EEDS, (c) output of the deep network of EEDS, (d) final result of EEDS.}
\label{fig:flow-result}
\end{figure}

{\flushleft \textbf{Discussion.} The performance of the deep and shallow networks in EEDS is reminiscent of prior residual prediction based methods~\cite{zeyde2010single,timofte2013anchored,timofte2014a+}, where SR is conducted by learning the residual between HR image and LR image upscaled by bicubic. As opposed to these approaches, our EEDS method replaces the fixed bicubic interpolation with a shallow network and jointly trains the deep and shallow networks, making the residual prediction based method a special case of our method.
From another perspective, the shallow network can also be interpreted as the shortcut connection proposed in the deep residual network~\cite{he2016}. However, the shortcut connection in \cite{he2016} is designed to facilitate gradient flows, while our shallow network is used to learn the major components of HR images and ease the difficulty of training the deep network.


%
\begin{flushleft} \textbf{Upsampling Analysis.}
To investigate the contribution of the upsampling module to the final results, we compare EEDS with three variants: EED (End-to-End Deep network), EED-ND (EED with no deconvolution), and EEDS-ND (EEDS with no deconvlution).
The EED-ND model is obtained by substituting the deconvolution layer of the deep network with a convolution layer producing the same number of channels.
%
Similarly, the EEDS-ND model is obtained by replacing the deconvolution layers of both deep and shallow networks in EEDS with convolution layers.
%
Correspondingly, both EED-ND and EEDS-ND take as input the LR images that have been upscaled to the desired sizes by bicubic interpolation.
%
\end{flushleft}

Tab.~\ref{tab:noDeconv} reports the average PSNR of the compared methods on three test sets, EEDS and EED considerably improves the performance of EEDS-ND and EED-ND, respectively, confirming that our learning based upsampling strategy in an appropriate feature space is more effective than directly pre-upscaling the LR image by bicubic interpolation in the original color space.

Furthermore, since the key parameter in the upsampling module is the kernel size of the deconvolution layer, additional evaluations are also conducted to study the performances of different kernel sizes.
While keeping the basic settings unchanged, we only modify the kernel size of deconvolution layer from the default value 15 to 7, 21 and 25, and denote their corresponding performances as EEDS-D7, EEDS-D21 and EEDS-D25, respectively.

Results in Tab.~\ref{tab:deconvKernel} show that the performance can be further improved by increasing of the kernel size, which suggests that the contextual information is beneficial for the task of upscaling. However, large kernel sizes entail more computational overhead. For both efficiency and effectiveness, we choose the size of 15 as a trade-off.

\begin{table}[htbp]
\linespread{1}
\centering
\caption{Average PSNR ($\mathrm{dB}$) of different upsampling strategies on Set5 with an upscaling factor 3.}
\label{tab:noDeconv}
\renewcommand\arraystretch{1.3}
\renewcommand{\multirowsetup}{\centering}
\begin{tabular}{|c|c|c|c|c|}
\hline
Model & EED-ND & EED & EEDS-ND & EEDS\\
\hline
Set5 & 33.01&33.20 & 33.22  & 33.47 \\
Set14 & 29.35& 29.46& 29.47 & 29.60\\
BSD100 & 28.46& 28.53 & 28.51  &28.64 \\
\hline
\end{tabular}
\end{table}

\begin{table}[htbp]
\linespread{1}
\centering
\caption{Average PSNR ($\mathrm{dB}$) of different kernel sizes for deconvolution layer on Set5 with an upscaling factor 3.}
\label{tab:deconvKernel}
\renewcommand\arraystretch{1.3}
\renewcommand{\multirowsetup}{\centering}
\begin{tabular}{|c|c|c|c|c|}
\hline
Model & EEDS-D7  & EEDS-D21 & EEDS-D25 & EEDS\\
\hline
Set5 &33.45 & 33.47&33.48 & 33.47 \\
Set14 & 29.57& 29.61&29.61 & 29.60\\
BSD100 &  28.63& 28.64&28.64 &28.64 \\
\hline
\end{tabular}
\end{table}

\begin{flushleft} \textbf{Multi-scale Analysis.}
In the reconstruction module of our EEDS model, the multi-scale convolution layer consists of four scales (kernel sizes): 1, 3, 5, 7.
To verify the effect of the multi-scale strategy for the image SR task, we compare the proposed multi-scale EEDS model with variants using single-scale (denoted as SS) reconstruction modules. For fair comparisons, instead of removing convolution layers with different kernel sizes, we set all the kernel sizes of the reconstruction module into the same sizes: $1 \times 1$, $3 \times 3$, $5 \times 5$, and $7 \times 7$, and obtain four variants denoted as EEDS-SS1, EEDS-SS3, EEDS-SS5 and EEDS-SS7, respectively.
%
\end{flushleft}

%

The performances of each scale (EEDS-SS) and multi-scale (EEDS) are reported in Tab. \ref{tab:singleScale}, indicating that large scale has better performance than small scale, due to the fact that large patches contain more contextual information than small ones.
Moreover, when fusing the four scales together for reconstruction, EEDS improves the average PSNR on Set5 by 0.20dB than EEDS-SS7, which validates that combining both short and long range context information can significantly benefit the ill-posed detail recovery problem.
%

\begin{table}[htbp]
\linespread{1}
\scriptsize
\centering
\caption{Average PSNR ($\mathrm{dB}$) of different reconstruction models on Set5 with an upscaling factor 3.}
\label{tab:singleScale}
\renewcommand\arraystretch{1.3}
\renewcommand{\multirowsetup}{\centering}
\begin{tabular}{|c|c|c|c|c|c|}
\hline
Model & EEDS-SS1 & EEDS-SS3 & EEDS-SS5 & EEDS-SS7 & EEDS \\
\hline
Set5 & 33.14&33.18&33.23 &33.27 &33.47\\
Set14 & 29.46&29.48 &29.48 & 29.49&29.60\\
BSD100 & 28.52 &28.54& 28.55& 28.55&28.64\\
\hline
\end{tabular}
\end{table}

\section{Conclusion}
\label{sec:conclusion}
This paper presents a novel end-to-end deep learning based approach for single image super-resolution, which directly extracts feature from the original LR image, and learns to upscale the resolution in the latent feature space. The final reconstruction considers both short- and long-range context by employing a multi-scale convolution layer.
In order to accelerate convergence rate for the deep network, we propose to jointly train the deep network with a relative shallow network which mainly takes the responsibility of learning the major image component while the deep network can better learn the residual.
Extensive experiments show that the proposed method can deliver more superior performance than state-of-the-art methods.
In addition, in-depth ablation studies are conducted to investigate the contribution of different CNN architectures for the task of image SR, which provides empirical knowledge for future research.
\bibliographystyle{IEEEtran}
\bibliography{IEEEabrv,refs}

\begin{thebibliography}{10}
\providecommand{\url}[1]{#1}
\csname url@samestyle\endcsname
\providecommand{\newblock}{\relax}
\providecommand{\bibinfo}[2]{#2}
\providecommand{\BIBentrySTDinterwordspacing}{\spaceskip=0pt\relax}
\providecommand{\BIBentryALTinterwordstretchfactor}{4}
\providecommand{\BIBentryALTinterwordspacing}{\spaceskip=\fontdimen2\font plus
\BIBentryALTinterwordstretchfactor\fontdimen3\font minus
  \fontdimen4\font\relax}
\providecommand{\BIBforeignlanguage}[2]{{%
\expandafter\ifx\csname l@#1\endcsname\relax
\typeout{** WARNING: IEEEtran.bst: No hyphenation pattern has been}%
\typeout{** loaded for the language `#1'. Using the pattern for}%
\typeout{** the default language instead.}%
\else
\language=\csname l@#1\endcsname
\fi
#2}}
\providecommand{\BIBdecl}{\relax}
\BIBdecl

\bibitem{glasner2009super}
D.~Glasner, S.~Bagon, and M.~Irani, ``Super-resolution from a single image,''
  in \emph{Proceedings of IEEE International Conference on Computer Vision},
  2009, pp. 349--356.

\bibitem{freedman2011image}
G.~Freedman and R.~Fattal, ``Image and video upscaling from local
  self-examples,'' \emph{ACM Transactions on Graphics}, vol.~30, no.~2, p.~12,
  2011.

\bibitem{yang2013fast}
C.-Y. Yang and M.-H. Yang, ``Fast direct super-resolution by simple
  functions,'' in \emph{Proceedings of IEEE International Conference on
  Computer Vision}, 2013, pp. 561--568.

\bibitem{freeman2000learning}
W.~T. Freeman, E.~C. Pasztor, and O.~T. Carmichael, ``Learning low-level
  vision,'' \emph{International Journal of Computer Vision}, vol.~40, no.~1,
  pp. 25--47, 2000.

\bibitem{freeman2002example}
W.~T. Freeman, T.~R. Jones, and E.~C. Pasztor, ``Example-based
  super-resolution,'' \emph{IEEE Computer Graphics and Applications}, vol.~22,
  no.~2, pp. 56--65, 2002.

\bibitem{yang2010image}
J.~Yang, J.~Wright, T.~S. Huang, and Y.~Ma, ``Image super-resolution via sparse
  representation,'' \emph{IEEE Transactions on Image Processing}, vol.~19,
  no.~11, pp. 2861--2873, 2010.

\bibitem{zeyde2010single}
R.~Zeyde, M.~Elad, and M.~Protter, ``On single image scale-up using
  sparse-representations,'' in \emph{Curves and Surfaces}, 2010, pp. 711--730.

\bibitem{yang2012coupled}
J.~Yang, Z.~Wang, Z.~Lin, S.~Cohen, and T.~Huang, ``Coupled dictionary training
  for image super-resolution,'' \emph{IEEE Transactions on Image Processing},
  vol.~21, no.~8, pp. 3467--3478, 2012.

\bibitem{chang2004super}
H.~Chang, D.-Y. Yeung, and Y.~Xiong, ``Super-resolution through neighbor
  embedding,'' in \emph{Proceedings of IEEE Conference on Computer Vision and
  Pattern Recognition}, vol.~1, 2004, pp. 275--282.

\bibitem{timofte2013anchored}
R.~Timofte, V.~Smet, and L.~Gool, ``Anchored neighborhood regression for fast
  example-based super-resolution,'' in \emph{Proceedings of IEEE International
  Conference on Computer Vision}, 2013, pp. 1920--1927.

\bibitem{timofte2014a+}
R.~Timofte, V.~De~Smet, and L.~Van~Gool, ``A+: Adjusted anchored neighborhood
  regression for fast super-resolution,'' in \emph{Proceedings of Asian
  Conference on Computer Vision}, 2014, pp. 111--126.

\bibitem{Schulter2015forests}
S.~Schulter, C.~Lesistner, and H.~Bischof, ``Fast and accutate image upscaling
  with super-resulution forests,'' in \emph{Proceedings of IEEE Conference on
  Computer Vision and Pattern Recognition}, June 2015, pp. 184--199.

\bibitem{huang2015fast}
J.~J. Huang, W.~C. Siu, and T.~R. Liu, ``Fast image interpolation via random
  forests,'' \emph{IEEE Transactions on Image Processing}, vol.~24, no.~10, pp.
  3232--3245, 2015.

\bibitem{Salvador2015naive}
J.~Salvador and E.~PerezPellitero, ``Naive bayes super-resulution forest,'' in
  \emph{Proceedings of IEEE International Conference on Computer Vision}, 2015,
  pp. 325--333.

\bibitem{dong2014learning}
C.~Dong, C.~C. Loy, K.~He, and X.~Tang, ``Learning a deep convolutional network
  for image super-resolution,'' in \emph{Proceedings of European Conference on
  Computer Vision}, 2014, pp. 184--199.

\bibitem{dong2016image}
------, ``Image super-resolution using deep convolutional networks,''
  \emph{IEEE Transactions on Pattern Analysis and Machine Intelligence},
  vol.~38, no.~2, pp. 295--307, 2016.

\bibitem{osendorfer2014image}
C.~Osendorfer, H.~Soyer, and P.~van~der Smagt, ``Image super-resolution with
  fast approximate convolutional sparse coding,'' in \emph{Neural Information
  Processing}, 2014, pp. 250--257.

\bibitem{wang2015deep}
Z.~Wang, D.~Liu, J.~Yang, W.~Han, and T.~Huang, ``Deep networks for image
  super-resolution with sparse prior,'' in \emph{Proceedings of IEEE
  International Conference on Computer Vision}, 2015, pp. 370--378.

\bibitem{gu2015convolutional}
S.~Gu, W.~Zuo, Q.~Xie, D.~Meng, X.~Feng, and L.~Zhang, ``Convolutional sparse
  coding for image super-resolution,'' in \emph{Proceedings of IEEE
  International Conference on Computer Vision}, 2015, pp. 1823--1831.

\bibitem{Kim2016deeply}
J.~Kim, J.~Kwon~Lee, and K.~Mu~Lee, ``Deeply-recursive convolutional network
  for image super-resolution,'' in \emph{Proceedings of IEEE Conference on
  Computer Vision and Pattern Recognition}, June 2016.

\bibitem{Kim2016accurate}
------, ``Accurate image super-resolution using very deep convolutional
  networks,'' in \emph{Proceedings of IEEE Conference on Computer Vision and
  Pattern Recognition}, June 2016.

\bibitem{shi2016real}
W.~Shi, J.~Caballero, F.~Huszar, J.~Totz, A.~P. Aitken, R.~Bishop, D.~Rueckert,
  and Z.~Wang, ``Real-time single image and video super-resolution using an
  efficient sub-pixel convolutional neural network,'' in \emph{Proceedings of
  IEEE Conference on Computer Vision and Pattern Recognition}, 2016, pp.
  1874--1883.

\bibitem{glorot2010understanding}
X.~Glorot and Y.~Bengio, ``Understanding the difficulty of training deep
  feedforward neural networks.'' in \emph{Aistats}, vol.~9, 2010, pp. 249--256.

\bibitem{he2016}
K.~He, X.~Zhang, S.~Ren, and J.~Sun, ``Deep residual learning for image
  recognition,'' in \emph{Proceedings of IEEE Conference on Computer Vision and
  Pattern Recognition}, 2016.

\bibitem{keys1981cubic}
R.~G. Keys, ``Cubic convolution interpolation for digital image processing,''
  \emph{IEEE Transactions on Acoustics, Speech and Signal Processing}, vol.~29,
  no.~6, pp. 1153--1160, 1981.

\bibitem{duchon1979lanczos}
C.~E. Duchon, ``Lanczos filtering in one and two dimensions,'' \emph{Journal of
  Applied Meteorology}, vol.~18, no.~8, pp. 1016--1022, 1979.

\bibitem{sun2008image}
J.~Sun, J.~Sun, Z.~Xu, and H.-Y. Shum, ``Image super-resolution using gradient
  profile prior,'' in \emph{Proceedings of IEEE Conference on Computer Vision
  and Pattern Recognition}, 2008, pp. 1--8.

\bibitem{protter2009generalizing}
M.~Protter, M.~Elad, H.~Takeda, and P.~Milanfar, ``Generalizing the
  nonlocal-means to super-resolution reconstruction,'' \emph{IEEE Transactions
  on Image Processing}, vol.~18, no.~1, pp. 36--51, 2009.

\bibitem{zhang2012single}
K.~Zhang, X.~Gao, D.~Tao, and X.~Li, ``Single image super-resolution with
  non-local means and steering kernel regression,'' \emph{IEEE Transactions on
  Image Processing}, vol.~21, no.~11, pp. 4544--4556, 2012.

\bibitem{bevilacqua2012low}
M.~Bevilacqua, A.~Roumy, C.~Guillemot, and M.-L. Alberi-Morel, ``Low-complexity
  single-image super-resolution based on nonnegative neighbor embedding,''
  2012, pp. 1--10.

\bibitem{lazebnik2006beyond}
S.~Lazebnik, C.~Schmid, and J.~Ponce, ``Beyond bags of features: Spatial
  pyramid matching for recognizing natural scene categories,'' in
  \emph{Proceedings of IEEE Conference on Computer Vision and Pattern
  Recognition}, vol.~2, 2006, pp. 2169--2178.

\bibitem{he2014spatial}
K.~He, X.~Zhang, S.~Ren, and J.~Sun, ``Spatial pyramid pooling in deep
  convolutional networks for visual recognition,'' in \emph{Proceedings of
  European Conference on Computer Vision}, 2014, pp. 346--361.

\bibitem{szegedy2015going}
C.~Szegedy, W.~Liu, Y.~Jia, P.~Sermanet, S.~Reed, D.~Anguelov, D.~Erhan,
  V.~Vanhoucke, and A.~Rabinovich, ``Going deeper with convolutions,'' in
  \emph{Proceedings of IEEE Conference on Computer Vision and Pattern
  Recognition}, 2015, pp. 1--9.

\bibitem{jia2014caffe}
Y.~Jia, E.~Shelhamer, J.~Donahue, S.~Karayev, J.~Long, R.~Girshick,
  S.~Guadarrama, and T.~Darrell, ``Caffe: Convolutional architecture for fast
  feature embedding,'' in \emph{Proceedings of the ACM International Conference
  on Multimedia}, 2014, pp. 675--678.

\bibitem{martin2001database}
D.~Martin, C.~Fowlkes, D.~Tal, and J.~Malik, ``A database of human segmented
  natural images and its application to evaluating segmentation algorithms and
  measuring ecological statistics,'' in \emph{Proceedings of IEEE International
  Conference on Computer Vision}, vol.~2, 2001, pp. 416--423.

\bibitem{wang2004image}
Z.~Wang, A.~C. Bovik, H.~R. Sheikh, and E.~P. Simoncelli, ``Image quality
  assessment: from error visibility to structural similarity,'' \emph{IEEE
  Transactions on Image Processing}, vol.~13, no.~4, pp. 600--612, 2004.

\bibitem{imagenet09}
J.~Deng, W.~Dong, R.~Socher, L.-J. Li, K.~Li, and L.~Fei-Fei, ``Imagenet: A
  large-scale hierarchical image database,'' in \emph{Proceedings of IEEE
  Conference on Computer Vision and Pattern Recognition}, 2009, pp. 248--255.

\end{thebibliography}

\end{document}